# Weakly supervised framework for wildlife detection and counting in challenging Arctic environments: a case study on caribou *(Rangifer tarandus)*

**Ghazaleh Serati[1,2], Samuel Foucher[1], Jérôme Théau[1,2]**

[1] Department of Applied Geomatics, Université de Sherbrooke, Sherbrooke, QC, Canada

[2] Quebec Centre for Biodiversity Science (QCBS), Stewart Biology, McGill University, Montréal Québec, QC, Canada

**Abstract**

Caribou across the Arctic has declined in recent decades, motivating scalable and accurate monitoring approaches to guide evidence-based conservation actions and policy decisions. Manual interpretation from this imagery is labor-intensive and error-prone, underscoring the need for automatic and reliable detection across varying scenes. Yet, such automatic detection is challenging due to severe background heterogeneity, dominant empty terrain (class imbalance), small or occluded targets, and wide variation in density and scale. To make the detection model (HerdNet) more robust to these challenges, a weakly supervised patch-level pretraining based on a detection network's architecture is proposed. The detection dataset includes five caribou herds distributed across Alaska. By learning from empty vs. non-empty labels in this dataset, the approach produces early weakly supervised knowledge for enhanced detection compared to HerdNet, which is initialized from generic weights. Accordingly, the patch-based pretrain network attained high accuracy on multi-herd imagery (2017) and on an independent year's (2019) test sets (F1: 93.7%/92.6%, respectively), enabling reliable mapping of regions containing animals to facilitate manual counting on large aerial imagery. Transferred to detection, initialization from weakly supervised pretraining yielded consistent gains over ImageNet weights on both positive patches (F1: 92.6%/93.5% vs. 89.3%/88.6%), and full-image counting (F1: 95.5%/93.3% vs. 91.5%/90.4%). Remaining limitations are false positives from animal-like background clutter and false negatives related to low animal density occlusions. Overall, pretraining on coarse labels prior to detection makes it possible to rely on weakly-supervised pretrained weights even when labeled data are limited, achieving results comparable to generic-weight initialization.

## 1. Introduction

In the Arctic, climate change is advancing more than twice as fast as the global average (Abrahms et al., 2023), leading to habitat loss, shifts in vegetation, reduced forage availability, and significant declines in terrestrial wildlife populations (Ito et al., 2020; Vynne et al., 2021). Among the diverse fauna inhabiting the circumpolar regions, caribou *(Rangifer tarandus)* is the most socio-ecologically significant terrestrial species, experiencing drastic population declines in recent decades (Joly et al., 2021; Lamb et al., 2024). The ever-increasing environmental changes, coupled with pronounced population declines in recent years, necessitate precise and frequent monitoring of their distribution and population (Cagnacci et al., 2016; Mallory & Boyce, 2018). However, caribou exhibit highly dynamic spatial behavior in response to environmental stresses. They form dense clusters during insect harassment and disperse across vast ranges during migration, further complicating consistent detection and monitoring (Lenzi et al., 2023a; Russell et al., 2021). These challenges are amplified by the Arctic's complex terrain, which includes diverse land cover such as shrublands, wetlands, forests, and snowfields. The variability of these environments introduces frequent changes in color, texture, and contrast. As a result, caribou can visually blend into their surroundings or be confused with background features, leading to missed or incorrect detections during aerial surveys.

Given these challenges, various remote sensing (RS) data sources such as imagery acquired from drones, aircrafts, and satellites have been increasingly utilized to improve wildlife monitoring across large or inaccessible natural landscapes (Converse et al., 2024; Corcoran et al., 2021; Delplanque, et al., 2024a). Among these, aerial imagery has emerged as a particularly effective approach for monitoring widely dispersed wildlife populations over vast areas (Davis et al., 2022; Keeping et al., 2018). Traditionally, large mammal surveys using aerial imagery have relied on manual interpretation, a process that is labor-intensive, prone to observer errors, and difficult to scale across the vast number of images required for large-area monitoring (Corcoran et al., 2019; Delplanque, et al., 2023a). However, with the rapid development of artificial intelligence (AI) techniques, particularly deep learning (DL), there has been a shift towards automated detection methods (Xu et al., 2024). While AI methods offer significant advantages in large-scale monitoring, specifically for handling vast amounts of data, they still face limitations. Most AI approaches, particularly those based on DL, depend on large, annotated datasets for effective optimization, which are challenging to obtain (Buckland et al., 2023; Converse et al., 2024; Lyons et al., 2019; Marchowski, 2021).

Additional challenges, such as small animal sizes, partial occlusion, proximity to one another, and distortions due to camera perspective or motion blur complicate DL-based detection (May et al., 2024). Additionally, the overwhelming prevalence of empty regions compared to those containing animals creates a considerable class imbalance in the datasets (Ke et al., 2024). This imbalance exacerbates the number of false negatives (FNs), where sparse or occluded animals are overlooked. Meanwhile, false positives (FPs) are often caused by background heterogeneity, where terrestrial features are misclassified as animals (Moreni et al., 2021a). Techniques such as data augmentation and weighted loss functions using hard negative patches (HNPs) have been employed to tackle this issue. Moreover, detection performance is further affected by camera perspective. While nadir imagery ensures consistent top-down representation, it lacks the contextual depth provided by oblique imagery (Delplanque et al., 2024b; Pashaei et al., 2020). Therefore, models that perform well on oblique imagery (Delplanque, et al., 2023a; Peng et al., 2020; Rančić et al., 2023) need to be tested on nadir imagery to ensure reliable performance across different perspectives (May et al., 2025).

In addition to these challenges, DL models in ecological monitoring often struggle to generalize across the complex, sparse, and heterogeneous Arctic landscapes (Bothmann et al., 2023). To address this limitation and enhance robustness, these models are typically initialized with ImageNet-pretrained weights, which provide general visual features and speed up fine-tuning on smaller, target datasets (Risojević & Stojnić, 2021). However, a key limitation is that ImageNet pretraining focuses on proximal ground-view imagery and may not transfer well to the detection of caribou in heterogeneous Arctic landscapes, both when individuals are densely aggregated or they are visually subtle at low densities. Addressing this domain mismatch with fully supervised learning would require vast amounts of well-annotated aerial imagery specific to Arctic wildlife monitoring, but this is hampered by the fragmented and varied nature of ecological monitoring data, as well as the significant time and specialized knowledge needed to label even modest volumes (Otarashvili et al., 2024).

To address these challenges, weakly supervised pretraining offers a promising strategy by enabling models to learn target representations from coarse labels such as patch-level presence information during early feature learning (Zhu et al., 2023). This approach is particularly valuable in Arctic wildlife monitoring, where most images contain no animals and annotation costs are high.

Prior work in both ecological monitoring and RS has explored weak supervision to reduce annotation costs, rather than using coarse labels for early-stage feature learning to mitigate the domain mismatch found in models initialized from general pretraining weights. For example, approaches such as pseudo-label generation frameworks (H. Wang et al., 2021) and confidence-based supervision methods (Yang et al., 2023; Zhu et al., 2024) leverage weak supervision within the main detection network by directly integrating pseudo-labels and image-level cues into the training process. In the context of wildlife monitoring, weak supervision has been applied to close-range imagery such as camera trap data and underwater footage to more easily reject empty images (de la Rosa et al., 2023; Pochelu et al., 2022) or to generate pseudo bounding-boxes or point annotations from weak image-level labels to alleviate the annotation burden (Xie et al., 2025; Berg et al., 2022; Kellenberger et al., 2019). However, these methods are typically implemented as standalone classifiers and do not transfer learned features to improve detection or reduce domain mismatch.

To our knowledge, no prior work in either ecological monitoring or RS has used coarse labeled pretraining based on the main detection network architecture to simultaneously address domain mismatch and enhance detection performance in large-scale aerial wildlife imagery, while also developing early weakly supervised knowledge about patches likely to contain animals.

The objective of this study is to improve DL-based single-class animal detection in large-scale Arctic aerial imagery by addressing core challenges such as extensive background heterogeneity, severe class imbalance due to the predominance of empty regions, and substantial variation in animal densities, scales, and occlusions. To this end, this study explores whether weakly supervised pretraining with coarse patch-level labels can yield features adapted to detection task that improve detection performance compared to conventional initialization strategies. The study also investigates the potential of patch-level pretraining to enhance early feature learning, which could help the model better learn relevant features before proceeding to the detection phase. Additionally, the study evaluates the ability of the pretraining network to accurately distinguish between patches containing animals and those that are empty, which may inform its broader applicability in data filtering or annotation support. Furthermore, the approach's generalization capability was assessed across multiple caribou herds and under temporal variation by testing on a dataset acquired in a different year.

## 2. Methods

Figure 1 illustrates the overall methodological pipeline for caribou detection. The process began with data preparation, which included converting geographical annotations to pixel coordinates, patching the aerial images, and filtering out patches containing image margins that are devoid of caribou. In the next step, a patch-based pretrain network (PPN) was trained to classify patches as empty or non-empty, initialized either from random weights or ImageNet weights. Throughout the first training phase, HerdNet (Delplanque, et al., 2023a), a point-based detection network, was trained to detect caribou locations throughout the positive patches using different initialization scenarios. Consequently, HNPs were extracted after the inference and added to the second training phase. Finally, the model's performance was evaluated on full images in terms of detection and counting accuracy.

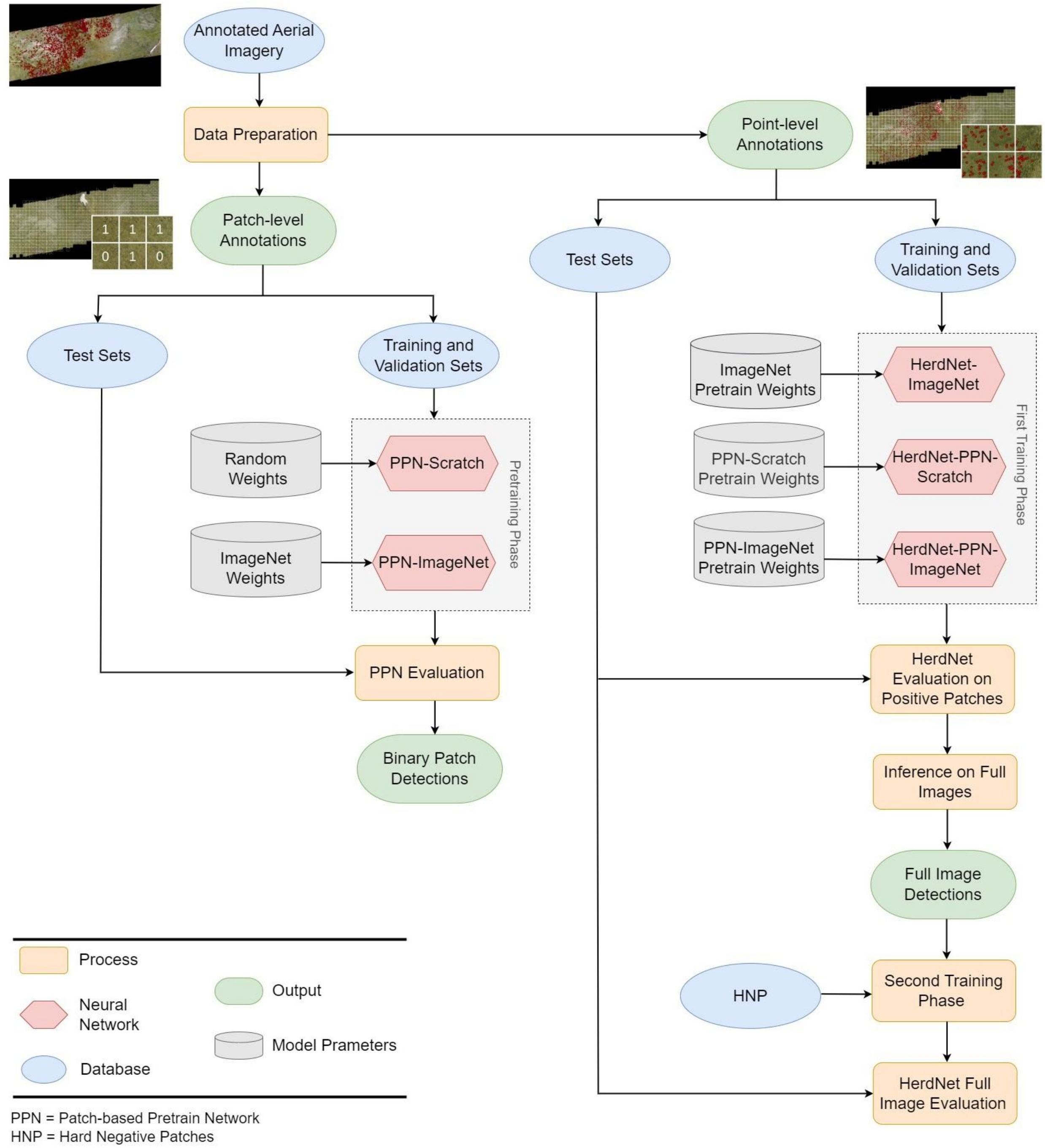

**Figure 1 :** Methodological pipeline for weakly-supervised caribou detection and its performance assessment within different initialization scenarios

### 2.1. Study Area and Dataset

The dataset represents a subset of a larger dataset acquired during extensive caribou surveys conducted by the Alaska Department of Fish and Game (ADF&G) over summer habitats. As illustrated in Figure 2, the study area spans northern and interior Alaska, encompassing a broad range of land cover types, ranging from tree cover in the boreal forest of interior areas to extensive grasslands, herbaceous wetlands, and moss-lichen dominated tundra in the northern regions. Patchy snow cover is also scattered across higher elevations and along the Arctic coastline. The surveyed regions include the contiguous ranges of five herds: the Central Arctic Herd (CAH), the Fortymile Caribou Herd (FCH), and the Porcupine Caribou Herd (PCH) in the northeastern and interior regions, and the distinct ranges of the Teshekpuk Caribou Herd (TCH) in the northwestern coastal plains and the Western Arctic Herd (WAH) across northwestern Alaska. Herd locations were identified using Global Navigation Satellite System (GNSS) collars, allowing discrete, targeted imagery acquisition over areas occupied by high-density caribou herds (Prichard et al., 2019a).

The dataset provided by the ADF&G included image mosaics from the five herds surveyed between July 1$^{st}$ and July 14$^{th}$, 2017, along with mosaics from the CAH dataset acquired on July 9$^{th}$, 2019. Each mosaic contained large images related to a specific scene, some of which contained no caribou. To ensure diversity and comprehensiveness in the dataset, five images per herd were manually selected from the 2017 dataset, prioritizing those with varying backgrounds and caribou densities. This selection focused on the diversity in land cover and animal density conditions, creating a diverse and robust dataset for DL-based caribou detection and generalization. Five images from the CAH 2019 dataset were also selected based on the same criteria to further evaluate the model's ability to generalize across a dataset acquired in a different year from distinct locations of the CAH's annual range. Figure 2 shows the location of selected aerial imagery, providing an overview of the landscape diversity captured in the dataset. Each point represents the center of the corresponding aerial image captured from that location.

The aerial imagery associated with each herd throughout the dataset encompasses distinct caribou spatial distributions and habitat preferences influenced by environmental conditions specific to each herd (Joly et al., 2010). At the time of imagery acquisition, the CAH's habitat was characterized by open tundra with patchy wetlands, where caribou exhibited moderate clustering near water bodies, and occasional construction sites. The FCH's habitat featured denser vegetation compared to other Alaskan caribou habitats, as well as scattered trees, with caribou gathering in open grassy areas or on snow patches. The PCH occupied mixed terrains, with expansive and dense clustering on snowfields and coastal areas, while the TCH's habitat consisted of a mix of wetlands, tundra, and small snow patches, with caribou gathering near water edges and permafrost areas, likely for cooling, insect relief, and foraging. Finally, the WAH was widely distributed over tundra, rivers, and barren rocky terrain. Both the 2017 and 2019 CAH datasets represent the same herd under broadly stable land-cover composition (Figure 2). However, 2019 imagery encompasses partially distinct locations, producing different background appearances that enable assessment of model generalization across years.

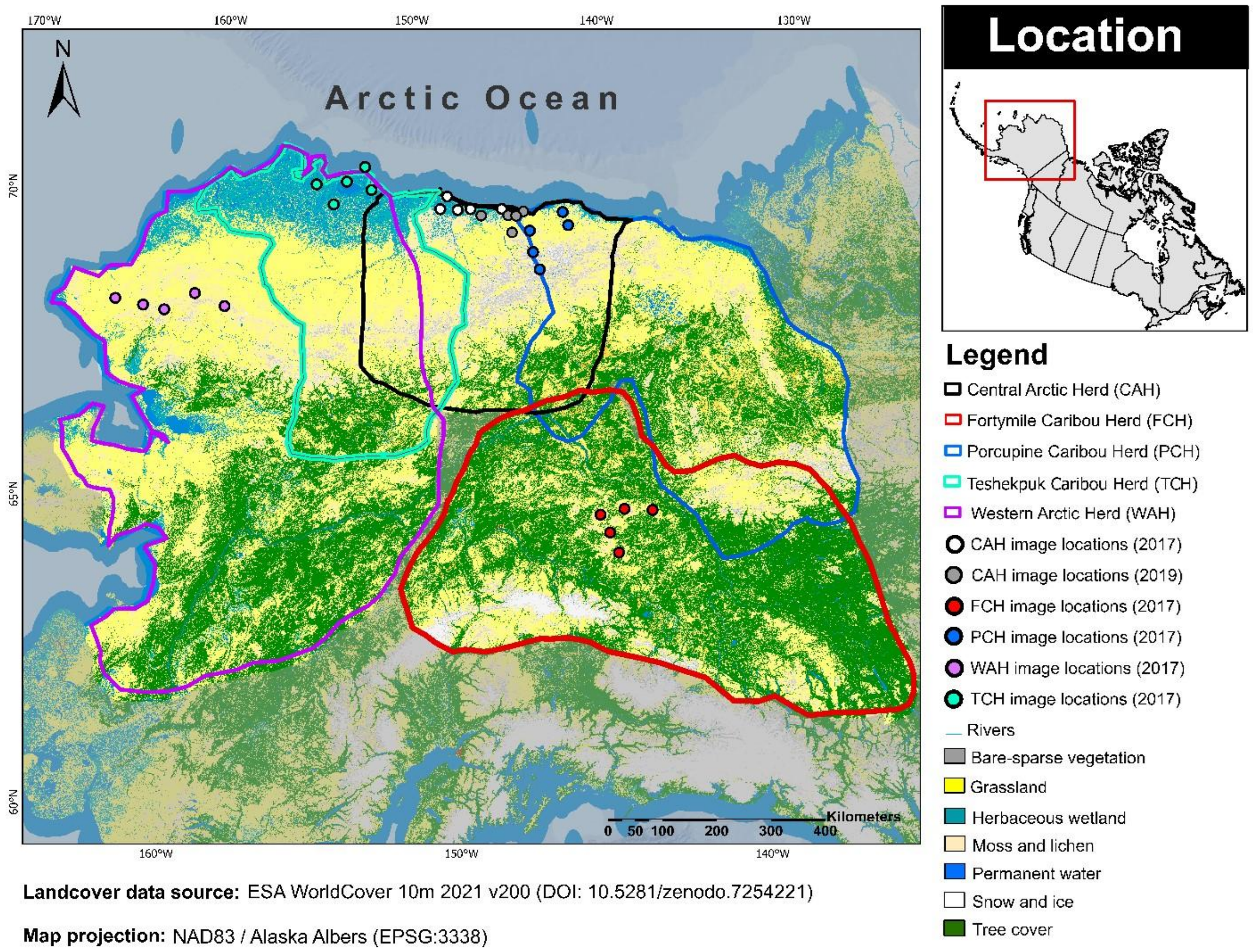

**Figure 2 :** Approximate annual ranges of understudy caribou herds in Alaska and the dominant landcovers in the area, as well as the locations of the selected images.

Table 1 provides an overview of the dataset, including ground sampling distance (GSD) values ranging from 2 to 4.3 cm, area of interest (AOI), and total caribou counts captured by the imagery associated with each herd within the selected images. By removing the margins lacking data, only image regions containing either background alone or background with caribou were used for training and evaluation of the models. For each herd, AOI is the aggregated ground footprint of the retained image regions. The dataset covers an approximate total AOI of 10.9 km$^2$. The number of caribou varied between herds, with certain herds (e.g. the WAH and the PCH) making a larger contribution to the overall dataset. Total caribou count represents the total annotation points for each herd, associated with their corresponding images in the dataset. Aerial imagery was acquired by three Phase One iXM-100 cameras (Phase One A/S, Copenhagen, Denmark) mounted on a DeHavilland Beaver aircraft, and the flight altitude was approximately 450 meters for both 2017 and 2019 surveys. Image sizes ranged from approximately 15,000 × 17,000 pixels to 25,000 × 25,000 pixels.

**Table 1 :** Summary of aerial image acquisition parameters and caribou counts across surveyed herds in the dataset

| Herd | Acquisition Year | GSD[1] (cm) | AOI[2] ($km^2$) | Total Caribou Count |
|---|---|---|---|---|
| CAH[3] | 2017 | 2 to 4.3 | 1.5 | 11,057 |
| | 2019 | 2.4 to 3.7 | 1.0 | 8,976 |
| FCH[4] | 2017 | 2.4 to 3.4 | 2.5 | 13,651 |
| PCH[5] | 2017 | 3 to 3.7 | 1.9 | 31,448 |
| TCH[6] | 2017 | 3.4 to 4 | 1.7 | 13,032 |
| WAH[7] | 2017 | 3.5 to 4.1 | 2.4 | 38,339 |

[1]GSD, Ground Sampling Distance; [2]AOI, Area of Interest; [3]CAH, Central Arctic Herd; [4]FCH, Fortymile Caribou Herd; [5]PCH, Porcupine Caribou Herd; [6]TCH, Teshekpuk Caribou Herd; [7]WAH, Western Arctic Herd.

## 2.2. Data Preparation

The 2017 images were allocated into training, validation, and test sets with a ratio of 70%, 10%, and 20%, respectively, ensuring that all five herds were represented across subsets based on point distributions. The allocation was done at the image level to avoid data leakage across subsets. The other test set, which involved imagery captured from CAH in 2019, covered more homogeneous backgrounds and fewer occlusions or environmental variations compared to the 2017 dataset. In contrast to the denser clusters observed in 2017, the caribou in the 2019 dataset were more sparsely distributed than in the 2017 dataset, with individuals spread across larger areas. Figure 3 presents image samples of the 2017 and 2019 datasets, featuring all the studied herds as well as various backgrounds, image dimensions and caribou distributions. Annotations are shown as red points.

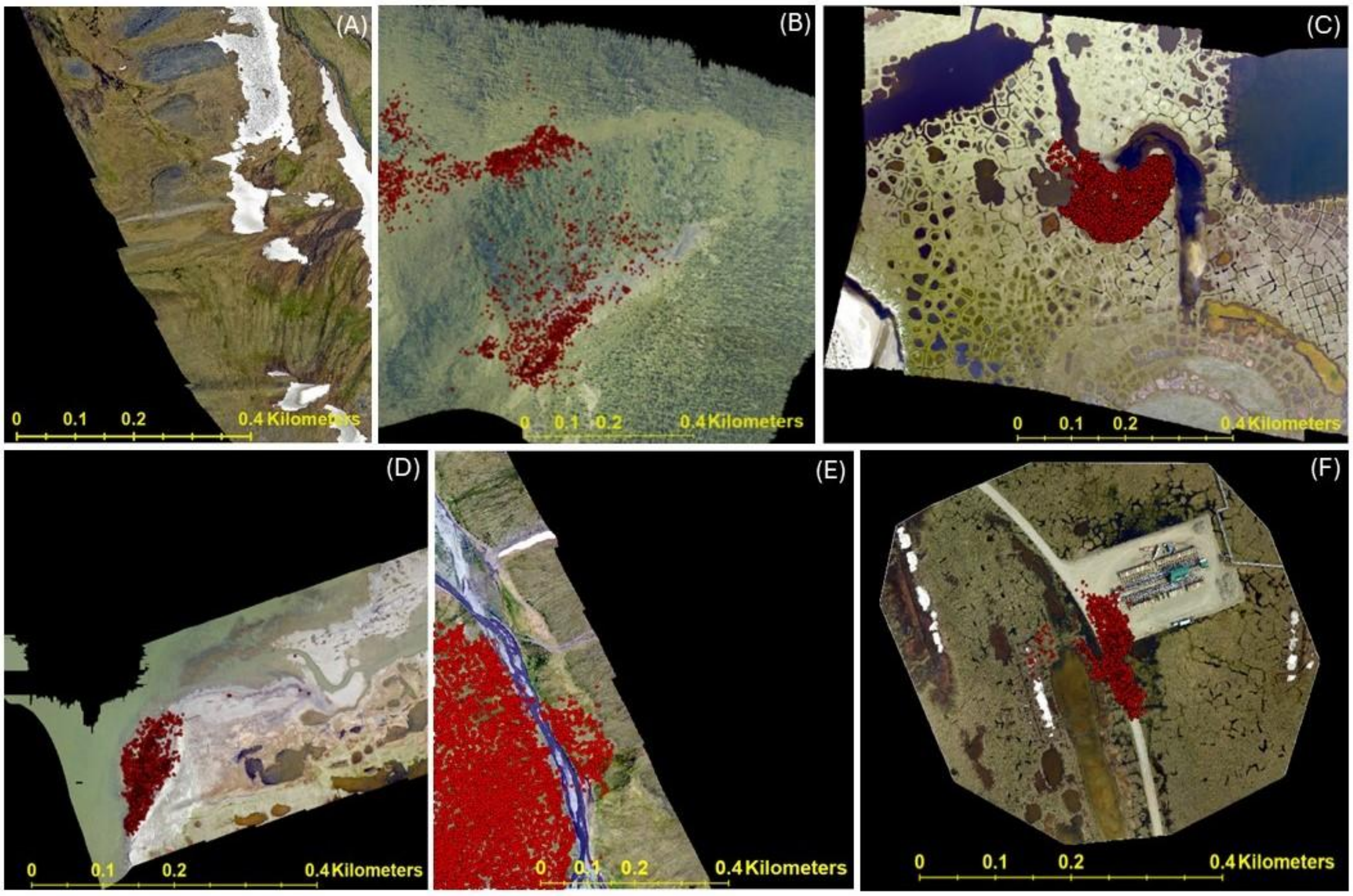

**Figure 3 :** Examples of images from different herds with varying backgrounds and caribou distribution patterns represented with red points; (A) Porcupine Caribou Herd (PCH); Background: snow, grassland, barren land; Density: high; (B) Fortymile Caribou Herd (FCH); Background: shrubland, trees, rocky terrain; Density: low to average; (C) Teshekpuk Caribou Herd (TCH); Background: wetland, grassland; Density: high; (D) Central Arctic Herd (CAH-2019); Background: frozen land, water; Density: average to high; (E) Western Arctic Herd (WAH); Background: water, grassland; Density: average to high; and (F) Central Arctic Herd (CAH-2017); Background: construction, grassland; Density: average to high.

The selected aerial images were then decomposed into small patches of 512 × 512 pixels to maintain high resolution while optimizing computational efficiency. Patch sizes in the range of 512 × 512 to 640 × 640 pixels have been shown to provide strong performance for small-object detection in aerial imagery (Unel et al., 2019) and have since become an established practice adopted in several animal detection studies. The training dataset consisted of 18,502 empty and 3,549 non-empty patches, with 9,294 empty and 1,181 non-empty patches in the 2017 test set and 3,650 empty and 1,025 non-empty patches in the 2019 test set. The animal sizes ranged from 5 to 14 pixels in length and 4 to 7 pixels in width, depending on age, image resolution, and visibility. A 15% patch overlap (around 78 pixels) was chosen to ensure that animals appearing near the edge of a patch would be included in at least one of the adjacent patches, preventing any individual from being missed. The overlap assured efficient capture of every individual for counting, while reducing redundancy and overfitting. Visual inspection confirmed that this overlap ensured complete animal inclusion across patch boundaries. Overall, this allocation strategy ensured that the model encountered a variety of patch-level densities, ranging from dispersed groups to occasional high-density clusters. This further enhanced model robustness across different environmental conditions and naturally variable caribou distributions. The distribution of annotated caribou points per patch across the training, validation, and test sets is available in Appendix 1.

To address class imbalance caused by a large number of empty patches, data stratification was applied to the validation data to maintain a consistent ratio of empty to non-empty patches across training and validation sets. For the 2017 dataset, the percentages of non-empty patches were 16% in the training and validation sets, and 11% in the test set, while this percentage was 22% in the 2019 test set. **Table *2*** illustrates the proportion of different herds in subsets of the 2017 dataset. This allocation ensured that all herds were represented in each data split (training, validation, and test), which enabled the model to learn from the diverse landscapes and caribou distributions associated with each herd, as the proportions were not numerically balanced due to the uneven distribution of caribou across herds in the dataset. Given the small number of images per herd, uneven caribou density, and the need to capture background variability within a fixed split, validation herd proportions differ from those in the training sets. Nonetheless, all herds are included in both splits.

**Table 2 :** Percentage of caribou associated with each herd across the 2017 dataset subsets, categorized by herds (%).

| Herd | Training | Validation | Test-2017 |
|---|---|---|---|
| CAH[1] | 7.4 | 7.4 | 33.3 |
| FCH[2] | 18.1 | 3.9 | 12.6 |
| PCH[3] | 38.2 | 25.0 | 33.3 |
| TCH[4] | 14.8 | 24.4 | 9.4 |
| WAH[5] | 21.5 | 39.3 | 11.4 |

[1]CAH, Central Arctic Herd; [2]FCH, Fortymile Caribou Herd; [3]PCH, Porcupine Caribou Herd; [4]TCH, Teshekpuk Caribou Herd; [5]WAH, Western Arctic Herd.

## 2.3. Deep Learning Approach

### 2.3.1. Patch-based Pretrain Network

Automatic wildlife detection over vast areas is particularly challenging due to the overwhelming presence of empty background patches in aerial imagery, which can lead to class imbalance, increased FPs, and ineffective feature learning (Azimi et al., 2018). Because most patches in aerial imagery are empty, the detection network struggles to learn meaningful representations for rare object locations. Additionally, occlusion and sparsity further increase the likelihood of FNs. To address these issues, and to provide weakly supervised feature extraction before the main detection task, a PPN was introduced. The PPN serves as a pre-training step that first learns to distinguish between empty and non-empty patches, allowing the model to learn features relevant to the dataset. By pre-training a detector backbone on this binary classification task, the detection network can more easily learn features related to the presence of animals, improving its ability to localize them while reducing FPs caused by background heterogeneity.

Throughout this paper, the term backbone is used to denote the entire encoder-decoder block of HerdNet, which is the down-sampling encoder together with its up-sampling decoder. The choice of using HerdNet's backbone for the PPN was based on its proven effectiveness in detecting individuals in dense herds where proximity and occlusion are common. Recent comparisons across point-based detectors also reported higher robustness of HerdNet in dense and occluded wildlife scenes compared to the point-based YOLO variant named POLO (May et al., 2025). Additionally, this approach allowed for an assessment of whether pretraining the backbone with a simple binary classification head and then transferring the adapted weights to initialize detection would improve HerdNet performance (Delplanque et al., 2023a). By utilizing the backbone of HerdNet in the PPN architecture, the model was pretrained on coarse binary patch labels (empty vs. non-empty) derived from the same dataset as the main detection task. This approach allowed the network to learn dataset-specific feature representations from these coarse labels, which were subsequently transferred to the main detection network's backbone, providing a more effective initialization for the subsequent detection task.

As shown in Figure 4, the PPN shares the same backbone as HerdNet, consisting of a deep layer aggregation (DLA) encoder to extract multi-level features, and a decoder that progressively reconstructs spatial information and generates feature maps at multiple scales (Yu et al., 2018). These feature maps highlight regions of interest within each patch. A spatial mean operation was then applied to summarize the feature map activations, and the resulting scalar was passed through a linear layer to produce the final binary classification output. A sigmoid activation function was applied to convert the output from the linear layer values into a probability, and this probability was subsequently thresholded to yield a final classification of 0 or 1. The network was trained using Binary Cross-Entropy (BCE) loss, where the predicted probability, obtained from the sigmoid activation, was compared to the ground truth label (0 for empty, 1 for non-empty).

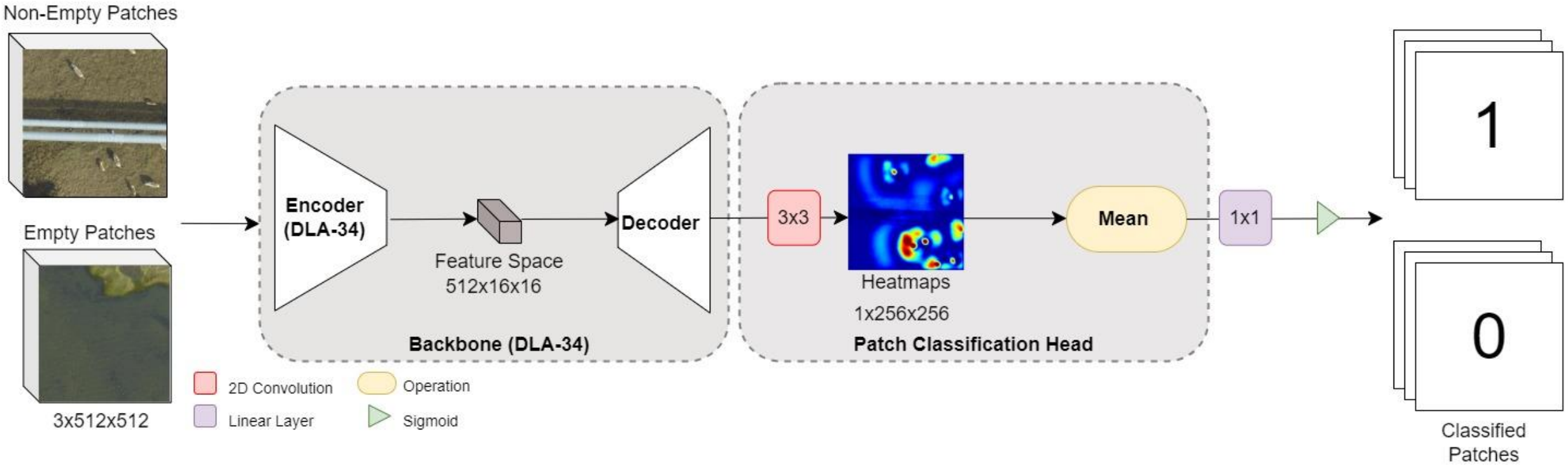

**Figure 4 :** Architecture of the patch-based pretrain network (PPN) using a deep layer aggregation (DLA-34) encoder-decoder backbone to classify patches as empty or non-empty.

During the training step, two models were trained. The first one involved initializing the weights of the DLA Encoder-Decoder using ImageNet pretrained weights (Russakovsky et al., 2015). The second training started from the PPN with a random weight initialization based on Xavier and Kaiming Normal distribution (Glorot & Bengio, 2010; He et al., 2015). For clarity, these two training approaches are called PPN-ImageNet and PPN-Scratch, respectively, throughout this paper.

Both training sessions were conducted under the same conditions, except for using a learning rate of $10^{-3}$ for initialization from random weights, compared to a lower rate of $10^{-4}$ for initialization from ImageNet weights. Consistent with standard fine-tuning practice, a lower learning rate was used for pretrained (ImageNet) weights to preserve useful features and avoid large, destabilizing updates, whereas a higher learning rate was required when training from scratch to accelerate early learning and convergence. The models were trained using the Adam optimizer, with a weight decay of $3\times10^{-4}$. To address the large class imbalance between empty and non-empty patches (~5:1 in the trained data), a Binary Batch Sampler (BBS) was used to ensure that each mini-batch contained an equal number of empty and non-empty patches during training. Empty patches were randomly selected in each epoch, while all non-empty patches were guaranteed to be used throughout the entire epoch. The total number of empty patches used was 18,502, and the total number of non-empty patches was 3,549. This method maintained a balanced class distribution in each batch, leading to a more effective model optimization and balanced feature learning compared to loss weighting, which could still introduce biases in learning (Singh et al., 2023). The BBS was applied only during training to prevent biases in metric calculation during validation and testing. Training was stopped after 152 epochs for PPN-ImageNet and 212 epochs for PPN-Scratch, when no further loss improvements were recorded. Early stopping was applied based on the validation loss, with a patience of 15 epochs.

Standard data augmentation techniques were applied to improve the robustness of the model. Binary patch-level annotations were generated automatically, based on the presence or absence of point annotations. These annotations were then manually refined to correct cases where patches containing parts of animals were incorrectly labeled as empty. To evaluate whether the improved features produced by the PPN led to better detection performance, the same backbone was applied within HerdNet, which is specifically designed for animal detection and counting. The next stage of the pipeline involved a detection network used to assess the impact of pretraining a model that shares the same backbone on the robustness of the detection network against challenging real-world detection scenarios.

### 2.3.2. Point-based Detection Network

HerdNet, a well-performing point-based detection network designed for the precise localization, classification, and counting of wild animals, was employed for the detection stage. The model has been shown to outperform Faster R-CNN and density-based networks in detection accuracy, counting precision, and robustness to dense herds, confirming its superiority in these tasks (Delplanque, et al., 2023a; Delplanque, et al., 2024b). HerdNet consists of a DLA-34 encoder to extract multi-scale hierarchical features, followed by a decoder that progressively upsamples feature maps. The network includes two output heads: a localization head, which predicts the focal instance distance transform (FIDT)-based heatmaps for pinpointing individual animal localization, and a classification head, which outputs low-resolution class maps to assign species labels to the detected points. Together, these output heads enable both accurate localization and species classification within dense and heterogeneous scenes.

Augmentation techniques similar to the PPN were applied to improve generalization. Throughout the first training step, the backbone weights were initialized using three different pretrained settings to evaluate the impact of various initialization strategies on the detection network's performance: 1) Weights derived from the PPN that were trained from scratch on the caribou detection dataset (HerdNet-PPN-Scratch); 2) Weights from the PPN, initialized with ImageNet weights (HerdNet-PPN-ImageNet); and 3) Backbone's pretrained weights on the ImageNet dataset (HerdNet-ImageNet), which is the baseline network in this study. All three scenarios were conducted using only the positive patches of the datasets. Early stopping was applied with a patience of 15 epochs to prevent overfitting. To evaluate the localization performance of the models throughout different initialization scenarios, patch-based animal counting was done for all the positive patches. The training process was conducted on an NVIDIA GeForce RTX 4090 GPU with 24GB of VRAM. The implementation was carried out using PyTorch, leveraging mixed-precision computation to optimize memory efficiency and speed. To manage memory constraints, a batch size of 32 was used for the PPN, while a batch size of 16 was allocated for training different HerdNet models due to its higher memory requirements.

In the second stage of training, HNPs were employed to address the challenge of reducing FPs in full-image detection (Hughes et al., 2018; Moreni et al., 2021a). Following the first training stage, which involved standard fine-tuning on positive patches, the second stage involved adding hard negative patches extracted from FP detections in full-image inference on training data. The number of HNPs added to the background class was 4,680, while the number of positive patches in the training dataset was 3,540, and the total number of training patches was 22,051. To mitigate the class imbalance between positive and negative patches, a BBS was also applied during the training. The second stage used the same patch size but applied a lower learning rate ($10^{-6}$), while other hyperparameters remained unchanged.

## 2.4. Model Evaluation

Each model was evaluated separately according to its intended function. The PPNs were evaluated based on their accuracy in classifying patches as empty or non-empty, and the point-based detection models were evaluated using detection and counting metrics.

### 2.4.1. Patch-based Pretrain Network Evaluation

In case of PPN, a prediction was assigned true positive (TP) when its patch-based ground truth label (0 or 1) matched the predicted value, and FPs were recorded for misclassified empty patches. Precision, recall, F1-score, and average precision (AP) were selected for patch-based evaluation. This evaluation determines how accurately the PPN maps animal presence in aerial patches and thus provides useful feature representations before detection network training.

### 2.4.2. Detection Network Evaluation

The evaluation of the detection network incorporated both localization accuracy and counting performance metrics, providing a complete analysis of its performance in localization and counting. The localization accuracy metrics include precision, recall, and F1 score. The counting performance metrics include mean absolute error (MAE), and total counting error (TCE), thereby assessing model accuracy in estimating the number of detected instances (Rančić et al., 2023; Zhang et al., 2024). A detection was labeled as a true positive (TP) when the Euclidean distance between it and the nearest ground-truth point was no more than 4 pixels, meaning that both points lay within a circle of 4-pixel radius. This threshold helps maintain spatial consistency between detections and ground truth points, minimizing over-counting in dense herds and enhancing localization precision.

# 3. Results

## 3.1. Pretrain Network Performance

The PPN was evaluated based on two initialization strategies: PPN-Scratch, trained from randomly initialized weights, and PPN-ImageNet, initialized using pretrained ImageNet weights. The performance of both models was assessed according to their ability to differentiate between empty and non-empty patches. Table 3 presents the results for both PPN-Scratch and PPN-ImageNet models using precision, recall, F1-score, and AP metrics, conducted on both 2017 and 2019 test sets. PPN-ImageNet consistently outperformed PPN-Scratch across all metrics throughout the two test sets. Even though the model had no prior exposure to the specific characteristics of the 2019 dataset, the ImageNet-pretrained model achieved slightly higher recall and comparable F1-score relative to the 2017 set, which highlights the model's robust performance across different years. Conversely, the PPN-Scratch model showed poor performance on both datasets, with similarly low recall and F1-scores, indicating overfitting to background patterns in the 2017 training data and failure to generalize to the 2019 imagery.

**Table 3 :** Patch-based pretrain network (PPN) results on the two test sets using different initialization strategies.

| Pretrain Network | PPN-ImageNet[1] | | PPN-Scratch[2] | |
|---|---|---|---|---|
| Test Set | 2017 | 2019 | 2017 | 2019 |
| Precision (%) | 92.2 | 89.1 | 85.4 | 83.5 |
| Recall (%) | 95.3 | 96.4 | 60.4 | 58.8 |
| F1-score (%) | 93.7 | 92.6 | 70.7 | 69.0 |
| AP (%)[3] | 93.8 | 91.5 | 66.2 | 63.7 |

[1]PPN-ImageNet, Patch-based Pretrain Network initialized from ImageNet weights; [2] PPN-Scratch, Patch-based Pretrain Network initialized from random weights; [3]AP, Average Precision.

Figure 5 shows some patch samples misclassified by the PPN-ImageNet as FPs (i.e., non-empty) throughout the two test sets. FPs in both test sets often include patches with background features such as rocks, logs, irregular ground textures, snow patches, water waves or other elements that share visual characteristics with lower-scale or partially occluded caribou individuals.

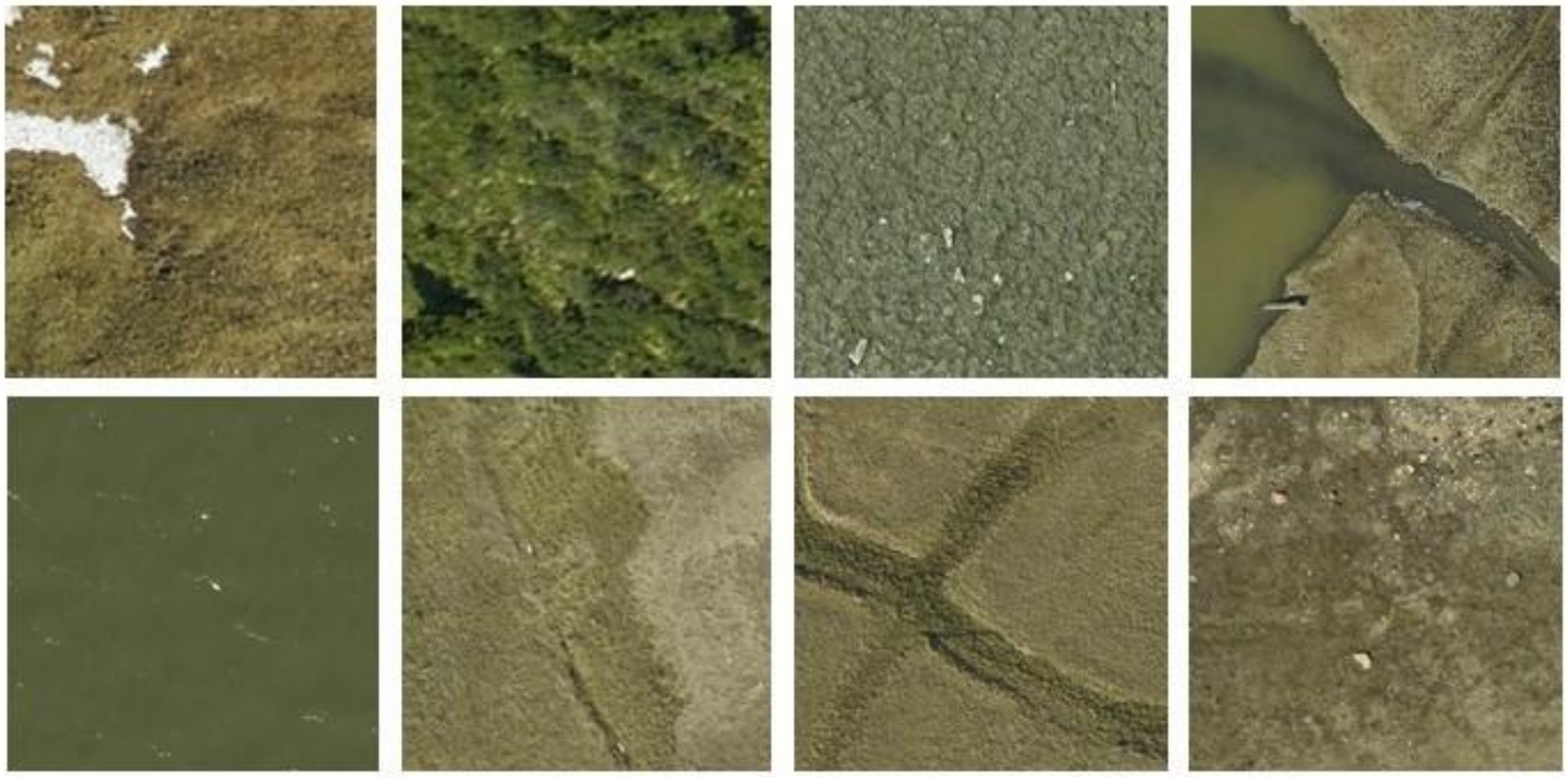

**Figure 5 :** Examples of false positive (FP) patches detected as non-empty by the patch-based pretrain network initialized from ImageNet weights (PPN-ImageNet). Top row: Examples from Test-2017; Bottom row: Examples from Test-2019.

On the other hand, FNs occurred mainly in the patches with low animal density. In 2017, all FN patches contained from 1 to 5 caribou, and the majority of the FNs (90.9%) were patches with one or two individuals, while FNs in the 2019 dataset involved 1 to 9 individuals, with 83.3% of all FNs containing one or two individuals. Regarding the FNs among all patches containing one or two caribou, the rates were respectively 23.0% and 16.6% in 2017, and 15.7% and 3.1% in 2019.

FN patches involved complex backgrounds, where caribou are mostly located near the patch margins, with cluttered scenes or lower contrast to the background (Appendix 2).

## 3.2. Animal Detection Network Performance

### 3.2.1. Detection Performance on Positive Patches

The performance of HerdNet under different initialization strategies was assessed after the first training phase. The evaluation was conducted on positive patches from the test sets, focusing on localization accuracy and patch-level counting performance. This approach avoided the confounding effects of severe class imbalance caused by empty background patches. The number of positive patches in the 2017 and 2019 test sets was 1,181 out of a total of 10,475, and 1,025 out of a total of 4,675, respectively. Table 4 summarizes the detection performance of three HerdNet initialization strategies evaluated on 2017 and 2019 test sets. Across all metrics, HerdNet-PPN-ImageNet consistently outperformed the other models, achieving the highest AP and F1-score values and lowest MAE on both test sets. However, HerdNet-PPN-Scratch achieved higher recall and slightly lower precision compared to the baseline HerdNet (HerdNet-ImageNet) introduced in Delplanque et al. (2023a) on both datasets. Notably, the results show that initializing the detection network with the weights generated from weakly supervised pretraining, whether initialized from ImageNet or random weights, noticeably increases the recall compared to initialization from generic weights.

**Table 4 :** Detection results of HerdNet under different initialization strategies on positive patch test sets.

| Model | HerdNet-PPN-Scratch[1] | | HerdNet-PPN-ImageNet[2] | | HerdNet-ImageNet[3] | |
|---|---|---|---|---|---|---|
| Test Set | 2017 | 2019 | 2017 | 2019 | 2017 | 2019 |
| Precision (%) | 88.6 | 88.5 | 92.5 | 91.1 | 91.2 | 90.4 |
| Recall (%) | 91.9 | 94.6 | 92.7 | 96.1 | 87.6 | 86.9 |
| F1-score (%) | 90.2 | 91.5 | 92.6 | 93.5 | 89.3 | 88.6 |
| MAE[4] | 1.3 | 1.0 | 0.8 | 0.7 | 1.2 | 1.1 |
| AP[5](%) | 85.3 | 89.8 | 92.3 | 91.6 | 81.4 | 88.6 |
| TCE[6](%) | 0.6 | 6.1 | 0.2 | 5.1 | -3.9 | -6.2 |

[1] HerdNet-PPN-Scratch, HerdNet initialized with weights from the patch-based network pretrained from scratch;
[2] HerdNet-PPN-ImageNet, HerdNet initialized with weights from the patch-based network pretrained on ImageNet;
[3]HerdNet-ImageNet, HerdNet initialized from ImageNet Weights (baseline); [4]MAE, Mean Absolute Error; [5]AP, Average Precision; [6]TCE, Total Counting Error.

Figure 6 and Figure 7 show patch-level count comparisons between ground truth (GT) and detections for the HerdNet-PPN-ImageNet and HerdNet-ImageNet models, on the Test-2017 and Test-2019 sets respectively. Each plotted point corresponds to a unique GT-Detection count pair. The color scale indicates the number of times each specific pair occurred, with brighter colors representing higher occurrences. Points closer to the diagonal line reflect more accurate predictions. A logarithmic scale is used on the axes to enhance visual separation, as patches with caribou counts between 0 and 10 occur far more frequently than those with higher counts. The $R^2$ coefficient shown in each scatterplot indicates how well the predicted values fit the GT values. Both figures reveal the higher alignment of HerdNet-PPN-ImageNet detections with GT counts compared to HerdNet-ImageNet, reflecting improved patch-based count prediction accuracy over the baseline HerdNet. Conversely, HerdNet-ImageNet undercounts the ground truth counts throughout the patches in both test sets, visible in Figure 6 and Figure 7 and as noted in Table 4 (TCE).

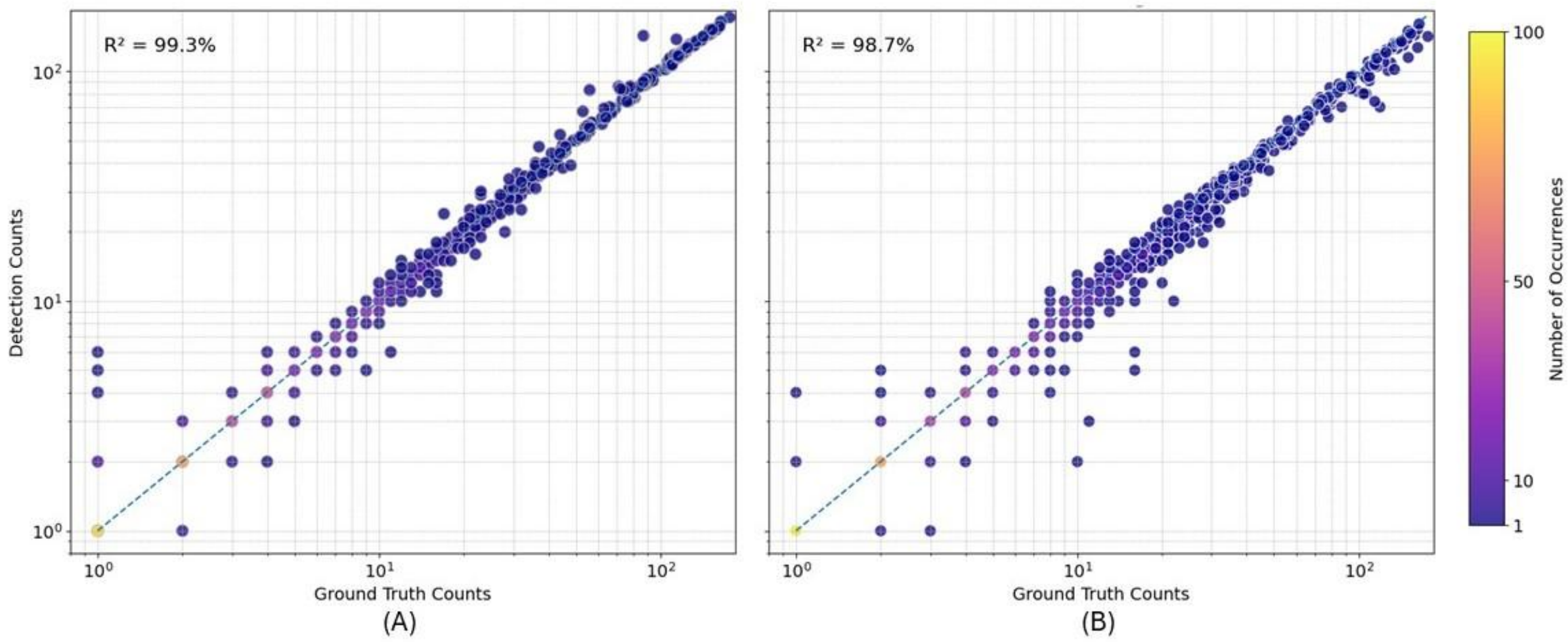

**Figure 6 :** Patch-level ground-truth vs detection counts on Test-2017: (A) HerdNet-PPN-ImageNet; (B) HerdNet-ImageNet (baseline).

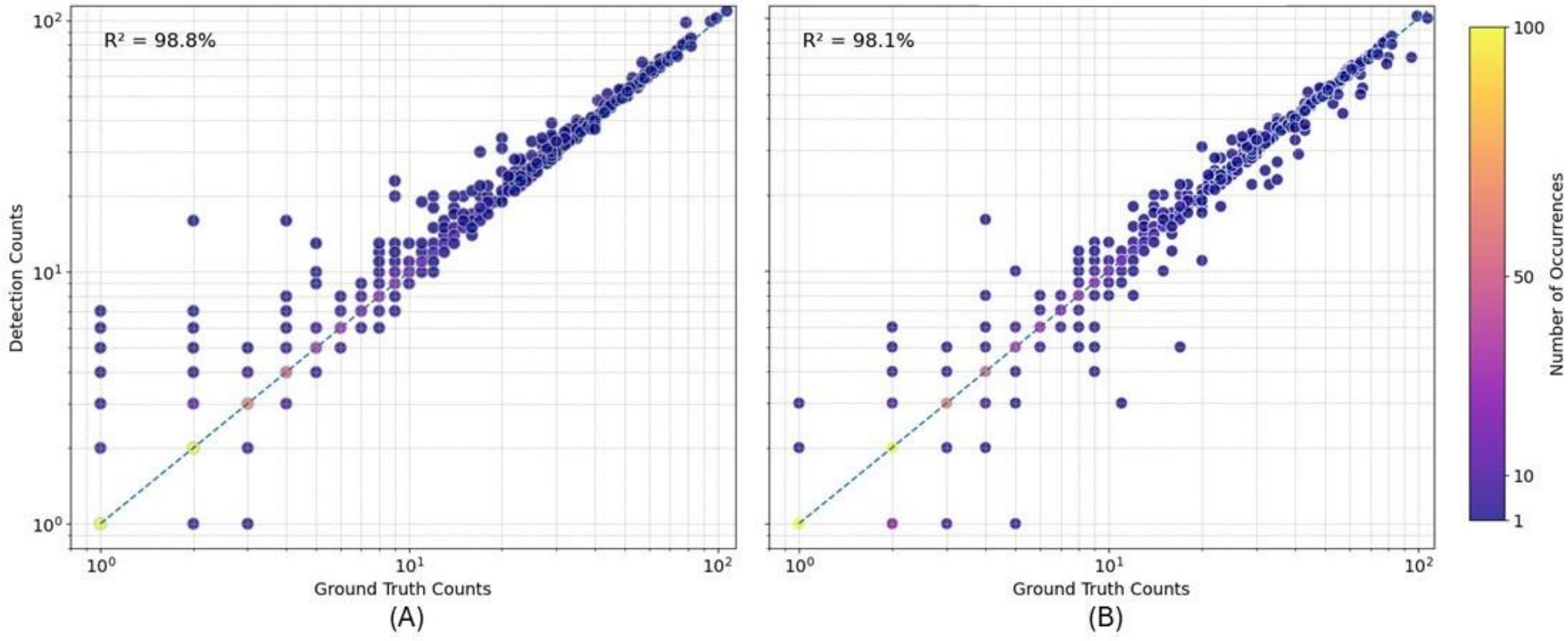

**Figure 7 :** Patch-level ground-truth vs detection counts on Test-2019: (A) HerdNet-PPN-ImageNet; (B) HerdNet-ImageNet (baseline).

### 3.2.2. Full Image Detection Performance

This section presents the full-image detection performance of the three HerdNet initialization variants after retraining them with a combination of positive patches and empty HNPs (2.3.2). Two examples of caribou detections in full-size aerial imagery from Test-2017 and Test-2019 are presented in Appendix 3. Table 5 reports key metrics on both the 2017 and 2019 test sets, evaluated on full images. The results show that all models benefited from HNPs, compared to the results of previous step in which the full image detections were retrieved from inference. HerdNet-PPN-ImageNet achieved the best overall performance across all localization and detection metrics. Due to the presence of overwhelming backgrounds, all HerdNet initializations tended to overestimate the total counts, reflected in positive TCE values. However, HerdNet-PPN-Scratch showed competitive performance with the baseline model (HerdNet-ImageNet), with lower precision causing higher TCEs, and improved recall in both datasets. Additionally, all models performed better on the 2017 dataset than on the 2019 dataset, likely due to the less challenging conditions and increased generalizability of the features learned, especially for models initialized with ImageNet weights.

**Table 5 :** Full image evaluation results of models resulted from different initialization scenarios after adding Hard Negative Patches (HNPs).

| Model | HerdNet-PPN-Scratch[1] | | HerdNet-PPN-ImageNet[2] | | HerdNet-ImageNet[3] | |
|---|---|---|---|---|---|---|
| Test Set | 2017 | 2019 | 2017 | 2019 | 2017 | 2019 |
| Precision (%) | 91.3 | 89.3 | 94.0 | 93.7 | 93.6 | 92.8 |
| Recall (%) | 94.5 | 90.8 | 97.2 | 93.0 | 89.6 | 88.3 |
| F1-score (%) | 92.8 | 90.0 | 95.5 | 93.3 | 91.5 | 90.4 |
| MAE[4] | 10.6 | 17.8 | 9.2 | 10.9 | 17.3 | 17.2 |
| AP[5] (%) | 94.3 | 90.0 | 96.6 | 92.8 | 91.1 | 90.3 |
| TCE[6] (%) | 0.5 | 6.7 | 0.1 | 3.1 | 0.3 | 3.4 |

[1] HerdNet-PPN-Scratch, HerdNet initialized with weights from the patch-based network pretrained from scratch;
[2] HerdNet-PPN-ImageNet, HerdNet initialized with weights from the patch-based network pretrained on ImageNet;
[3]HerdNet-ImageNet, HerdNet initialized from ImageNet Weights (baseline); [4]MAE, Mean Absolute Error;
[5]AP, Average Precision; [6]TCE, Total Counting Error.

The caribou detected in different challenging conditions using HerdNet-PPN-ImageNet are shown in Figure 8, which represents sample patches from the test sets. These samples demonstrate the model's ability to accurately localize caribou despite challenges such as high density (figure 8A), occlusions and shadows (Figures 8B and C), background variability and varying scales (Figures 8A-D), all present in real-world Arctic backgrounds. Moreover, HerdNet-PPN-ImageNet showed lower MAE with lower variability compared to HerdNet-ImageNet and HerdNet-PPN-Scratch (Appendix 4). Additionally, statistical significance of the observed improvements was confirmed using paired Wilcoxon signed-rank tests on per-patch counting errors (Wilcoxon, 1945). The resulting p-values were 7.5 × 10-25 for the Test-2017 dataset and 3.2 × 10-11 for the Test-2019 dataset, indicating highly consistent improvements across both survey years.

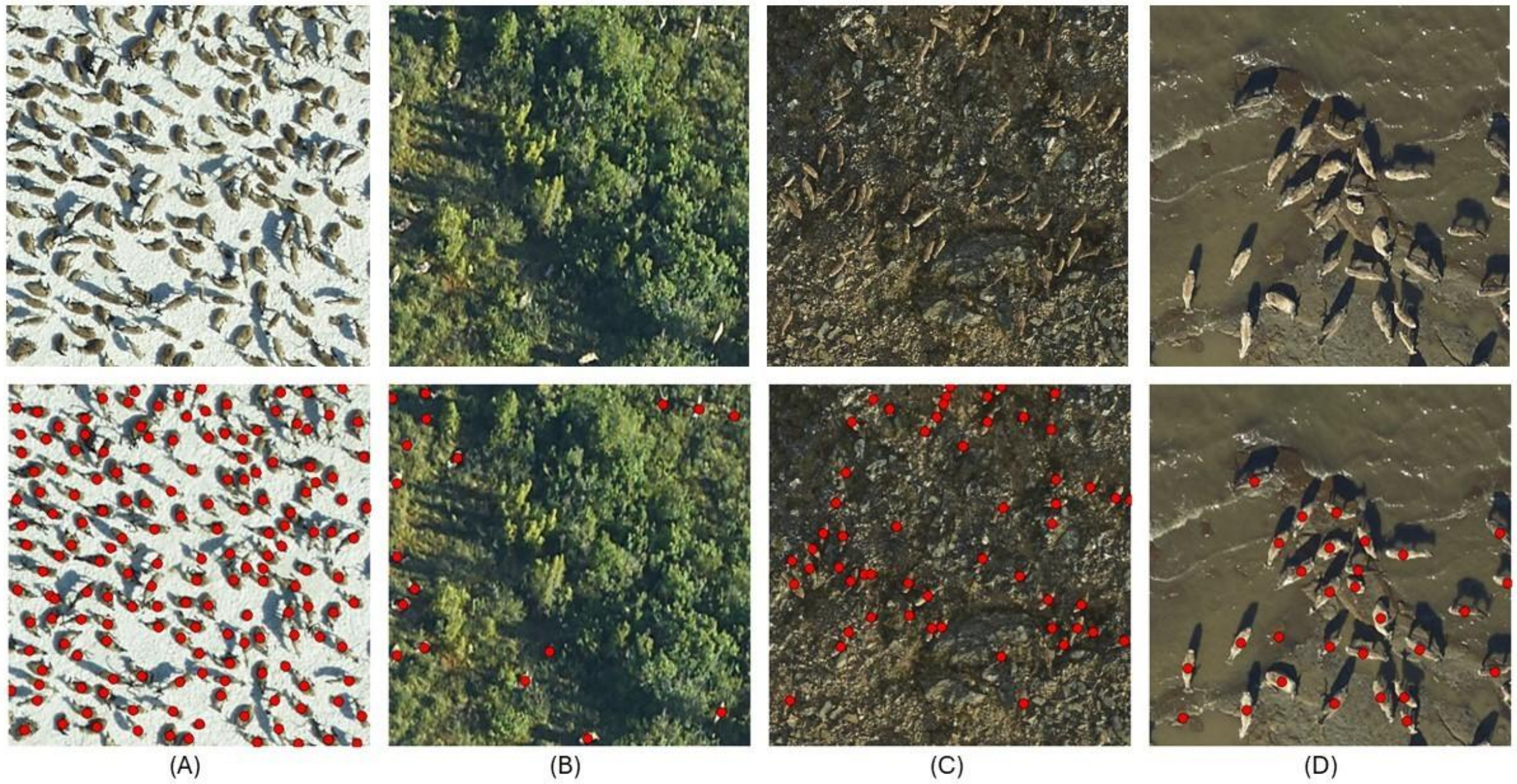

**Figure 8 :** Examples of HerdNet initialized with weights from the patch-based network pretrained on ImageNet (HerdNet-PPN-ImageNet) predictions under some challenging conditions; Top row: Patches from the 2017 test set; Bottom row: Corresponding detections represented by red points: (A) High caribou density; (B) Occlusions; (C) Shadow and heterogeneous background; (D) Higher spatial resolution compared to the majority of the images in the dataset.

## 4. Discussion

The visual complexity of natural environments and the sparse and uneven distribution of animals limit the effectiveness of DL-based detection models. This is particularly the case when relying solely on generic pretrained weights, which may not capture specific features relevant to animal detection tasks. This study demonstrated that pretraining the backbone of a detection network (HerdNet) on the same dataset as the detection task using weak patch-based annotations (i.e., empty vs. non-empty) improves the detection network's performance in tackling challenging conditions. This method also achieved competitive results even when ImageNet initialization was entirely omitted. These findings indicate the effectiveness of patch-based pretraining on a dataset before moving to the detection task on increasing the robustness of the detection network. Moreover, when initialized with ImageNet weights, the PPN serves as a pre-filtering tool to flag patches likely to contain caribou to focus and speed up the manual annotation process.

### 4.1. Performing Weakly Supervised Pretraining Prior to the Detection Task

Among the two initialization strategies applied to pretrain network, PPN-ImageNet was the best-performing pretraining network. It was obtained by fine-tuning the ImageNet-initialized HerdNet backbone with a simple binary patch-based head attached during pretraining. The high recall of PPN-ImageNet on the two sets further reduced the chance of missing patches containing animals. This enhancement is particularly valuable for semi-automated frameworks designed to filter out backgrounds. In such frameworks, prioritizing recall ensures that true animal detections are less likely to be missed and can be subsequently validated by human reviewers (Delplanque, et al., 2023a). While weak supervision has been explored in other ecological applications, prior work primarily focused on filtering full empty images in camera trap datasets using feature similarity learning rather than explicit weak labels (de la Rosa et al., 2023; Pochelu et al., 2022). Weak labels have also been used to reduce annotation burdens in DL-based animal detection tasks (Kellenberger et al., 2019), but not to improve detection accuracy under complex field conditions as demonstrated in this study.

Alongside these advantages, PPN-Scratch's lower precision and recall relative to PPN-ImageNet highlights the challenge of achieving reliable results when trained from scratch on a limited dataset. This approach accentuated the model's background bias in patch classification, despite the use of BBS during training to address class imbalance. Nevertheless, the observed detection performance gap between PPN-Scratch and PPN-ImageNet can be attributed to the fact that generic weights provide low-level and mid-level visual features that are still broadly transferable to RS and wildlife monitoring tasks. As a consequence, models initialized with generic weights achieve better performance than those starting from random initialization on relatively smaller datasets (Buschbacher et al., 2020; Chen et al., 2022; Corley et al., 2024). Regarding false alarms in PPN-ImageNet, visually confusing small background structures (e.g., fine textures, shadows, vegetation) triggered animal-like responses and produced FPs. Furthermore, sparsely occurring animals near patch boundaries as well as dispersed individuals partially occluded due to background heterogeneity or texture and color similarity, were missed as FNs, due to limited contextual evidence. These patterns were consistent with sources of errors previously reported for empty-non-empty camera-trap imagery screening models (Banerjee et al., 2022; Y. Wang et al., 2024). Although boundary-adjacent animals inevitably introduce some level of weak-label noise, their contribution to PPN derived FNs was found to be minimal. Patches containing low-density caribou (less than 5 and 9 individuals for 2017 and 2019 test sets respectively) within 10 pixels of a patch boundary accounted for only 1.3% (Test-2017) and 1.4% (Test-2019) of all non-empty patches, making them a very small subset of the candidate patches that could produce FNs. Within this small subset, FNs occurred primarily when animals were tightly clustered along a single patch edge in which limited receptive field context weakened the heatmap response. In contrast, patches where animals were more evenly distributed near the margins, even at low densities, rarely produced FNs. Given both the rarity of such boundary-sensitive cases and their concentration in low-density patches (1-3 individuals), their impact on the overall PPN metrics is negligible.

### 4.2. Effect of Weakly-supervised Pretraining on Detection Performance

Regarding the effect of weakly supervised pretraining on localization accuracy, incorporating a weakly-supervised patch-level pretraining enhanced both recall and precision, with recall having higher gains than precision. As a result, animal detection performance increased in challenging Arctic aerial imagery. This performance improvement was consistent over both positive patch-level and full image evaluations with the presence of empty regions. HerdNet-PPN-ImageNet achieved consistently superior performance across all detection and counting metrics compared to the base HerdNet, effectively mitigating the underestimation bias in the base HerdNet model (Delplanque, et al., 2023a). Notably, even HerdNet-PPN-Scratch with no reliance on generic ImageNet weights demonstrated higher recall and competitive precision compared to the baseline (HerdNet-ImageNet). The findings also demonstrated that pretraining a weakly supervised network extracted from the detector's backbone on the detection dataset, before going through the main detection procedure, could convert random initial weights into weights specifically adapted to the data characteristics. These pretrained weights allowed the model to capture subtle animal features in wide, complex backgrounds and to achieve detection performance comparable to that of a model initialized from generic pretrained weights. Similar to previous studies that have demonstrated the effectiveness of adding HNPs to the positive patches to reduce excessive FPs in wildlife detection, the current study also found that incorporating HNPs led to improved full-image detection performance across all HerdNet models.

Having a pretraining stage prior to the main detection task proposed in the current work targets a concept similar to self-supervised learning (SSL) strategies that have recently been applied to RS data (Bourcier et al., 2022; Guo et al., 2025; Manas et al., 2021). These approaches are primarily motivated by the limited availability of labeled RS data and aim to learn general features from large unlabeled satellite datasets (Muhtar et al., 2023). Although such in-domain SSL pretraining can reduce dependence on fully labeled data and improve the main network performance, it has been mainly explored for land-cover classification, semantic segmentation, or few-shot recognition, rather than object detection as the target task (Y. Wang et al., 2022). Moreover, these frameworks usually only pretrain the encoder (often ResNet or Vision Transformer variants) to learn general visual representations from unlabeled imagery. This process lacks a specific head that captures meaningful knowledge about the data distribution or the overwhelming prevalence of empty regions in RS imagery. In contrast, our weakly supervised pretraining explicitly updates both the encoder and the decoder within the same architecture as the detection model, yielding reliable results on the patches containing animals. This enables the transfer of richer and more task-aligned features, improving model robustness under challenging detection conditions.

Background heterogeneity is caused by the variation in land cover, texture, illumination, and scale that diminishes target-background separability in image-based animal detection problems. Results demonstrated that in addition to expanding the training data diversity, updating the detection network's backbone with a weakly supervised pretrain model improved HerdNet's robustness to this challenge, and reduced FNs. More specifically, despite recent advances, excessive FPs due to visual similarities between background features and animals (Peng et al., 2020), class imbalance and annotation burden (Kellenberger et al., 2019), missed detections of low-density animals (Roca et al., 2024), and limited scalability and generalization (Lenzi et al., 2023a) still remain the most common error sources in automated wildlife detection. Additionally, even specialized models such as HerdNet struggle to reliably distinguish minority species and to avoid underestimation in mixed or densely clustered herds over complex backgrounds. A key cause is the classification head's coarse spatial resolution (Delplanque, et al., 2023a). The proposed pipeline helped the classification and localization heads of HerdNet to acquire prior knowledge adapted to the detection task by updating the encoder and decoder weights in HerdNet architecture. This increased the robustness of HerdNet in distinguishing individuals within dense herds, as well as in handling low caribou density scenes, background heterogeneity, and class imbalance. As a result, the proposed initialization yielded more reliable detections in Arctic aerial imagery than the widely adopted practice of initializing from generic weights such as ImageNet or COCO (Roca et al., 2025; Delplanque, et al., 2023a; Lenzi et al., 2023; Moreni et al., 2021; Dominguez-Morales et al., 2021; Kellenberger et al., 2019; Moreni et al., 2023).

### 4.3. Insights into Large-Scale Caribou Counting for Wildlife Surveys

In Arctic ecosystems, where caribou are scattered across vast and complex landscapes, aerial surveys offer a practical, continuous, and non-invasive method for monitoring (Poole et al., 2013; Prichard et al., 2019b). However, the prevalence of empty regions and the heterogeneous, varying landscapes of Arctic summer habitats make manual counting time-consuming and challenging (Delplanque et al., 2023a; Terletzky & Ramsey, 2016). Integrating the proposed PPN would enable survey teams to quickly discard empty regions and focus on candidate areas within large aerial mosaics. With spatially-referenced patches, this pre-screening step would improve survey efficiency and markedly reduce manual interpretation time. Compared to the baseline (Delplanque et al., 2023a), HerdNet-PPN-ImageNet showed robust full-image counting performance, including on imagery collected from distinct herds in a different year than the training data. This multi-year robustness highlighted its potential for long-term and multi-herd Arctic caribou monitoring without the need for retraining on new data. Notably, this robustness was achieved using a relatively small, annotated dataset. Scaling up the volume of pretrained data would likely improve the model's adaptability and detection accuracy even further and omit the need for using generic weights to initialize the PPN.

To our knowledge, the only prior study on automated caribou detection is Lenzi et al. (2023). They performed multi-class detection of caribou by demographic categories (adult, and juvenile) on low-altitude UAV imagery collected over small study areas within protected caribou habitats in Canada, where backgrounds were relatively homogeneous. Nevertheless, challenges such as background clutter from vegetation hindered robust automatic multi-class caribou detection. Building on these insights, the present study improved the detection performance of HerdNet in large-scale Arctic regions characterized by heterogeneous habitats and substantial variation in caribou densities. The robustness was further validated on imagery collected in two survey years. In addition, it extended the scope of application to automated full-image animal counting scenarios with extensive empty areas, an ecologically important task that is still largely performed manually on aerial imagery. The model's ability to generalize across geographically diverse caribou habitats associated with different herds in the Arctic also indicated its potential applicability to other circumpolar regions, such as Canada, Scandinavia, and Russia caribou habitats (Cameron & Kennedy, 2023; Campbell et al., 2021; Davison, 2016; Gunn & Russell, 2022).

In dense wildlife aggregations, the choice of detection paradigm becomes critical. In anchor-based methods, bounding boxes associated with adjacent animals become highly overlapping. This can cause non-maximum suppression to suppress true positives or merge detections, ultimately leading to biased counts (Delplanque et al., 2022; Lenzi et al., 2023; Peng et al., 2020; Roca et al., 2024; Roy et al., 2023). In contrast, point-based detection avoids bounding-box overlap ambiguity, requires simpler annotations, and remains robust when individuals are tightly clustered, making it well-suited for dense aggregations. Caribou frequently form high-density aggregations during warmer seasons, with smaller individuals occurring near larger ones. To this end, a point-based detection model (HerdNet) is adopted and improved to conduct reliable detection and counting in Arctic aerial surveys.

The enhanced robustness of the HerdNet-PPN-ImageNet model across different herds and survey years, increases its suitability for annual wildlife monitoring over vast and diverse scenes. Since PPN reliably detects patches containing caribou, analysis can be concentrated on regions with caribou presence. This enables faster identification of herd locations and supports assessments of movement patterns between surveys. Combined with improved full-image point detections, this substantially reduces the manual counting efforts while maintaining the reliability on full-image counts. Nevertheless, human review is still necessary for final confirmation, while significantly reducing the manual workload. The combined animal detection and patch-filtering capabilities of HerdNet-PPN-ImageNet also suggest that its potential applicability extends beyond aerial surveys. The PPN can flag patches that may contain caribou while excluding many empty scenes. This capability could be valuable for very-high-resolution satellite imagery, where acquisitions are costly and most images contain no animals (Delplanque et al., 2024). Testing such an approach on satellite data would help determine whether it can reliably highlight informative regions and detect dense herds in the imagery that is costly to acquire and has lower spatial resolution than aerial images.

### 4.4. Toward Scalable and Adaptive Wildlife Detection

The demonstrated robustness of PPN-ImageNet compared to HerdNet-ImageNet suggests several promising directions for enhancing the proposed method's performance and applicability. One potential enhancement involves increasing the complexity of the pretraining task from simple object presence estimation to a more sophisticated pretraining task. For instance, this could involve estimating the number of animals per patch, enabling the network to capture both presence and spatial density. As a result, the backbone network could capture more enriched pretraining features and further improve the detection performance under highly variable conditions.

Due to the labor-intensive nature of annotating large-scale aerial imagery, the pretraining dataset was restricted to a limited number of images in the current study. Consequently, the network's exposure to diverse background and animal configurations was constrained, potentially limiting its generalizability across broader spatial variations. Thus, a practical approach could involve using the trained HerdNet-PPN-ImageNet model to perform inference on unannotated full images, generating pseudo-annotations for further training. These inferred results could then be manually verified and corrected to provide reliable annotations, which can be extended to future annual survey data. By incorporating this enriched dataset, the binary detection network trained from scratch would become more robust and learn more discriminative and data-driven features, allowing it to handle the complexities of large-scale wildlife monitoring tasks and reducing the need for ImageNet weights. While this study focused on single-class caribou detection, evaluating the effectiveness of this patch-based pretraining approach in multi-class detection settings would be an important direction. For instance, in the case of Lenzi et al. (2023), where the goal was to distinguish between juvenile and mature caribou, it would be valuable to evaluate how patch-level pretraining affects class separation and species-specific detection within HerdNet architecture.

## 5. Conclusion

Monitoring dynamic caribou populations across vast and visually complex Arctic environments presents unique challenges for automated detection models. Sparse animal distributions, dense herds containing thousands of animals, heterogeneous backgrounds, and high annotation costs make it difficult to build robust detection pipelines using traditional deep learning approaches. Furthermore, the dominance of empty regions within large-scale aerial imagery, where animals are sparsely and unevenly distributed, contributes to severe class imbalance. On the other hand, standard pretraining on generic image datasets like ImageNet fails to fully capture the specific characteristics of Arctic wildlife scenes.

In this study, a weakly supervised two-stage pipeline tailored for detecting caribou in Arctic aerial imagery was proposed. As part of this approach, the detection network's backbone was trained on the same dataset as the one used for detection, using a patch-based binary classification task. This allowed the network to be pretrained using weak labels, enabling it to reliably distinguish between empty and non-empty image patches. The weakly-supervised initialization likely improved the detection model's ability to capture features specific to the detection task, as opposed to the general features learned from ImageNet, leading to better performance. Combined with adding HNPs and evaluation across two temporally and visually distinct datasets, this approach demonstrated improved detection robustness, reduced FPs, and enhanced recall, highlighting its potential for large-scale wildlife monitoring in complex and dynamics environments like the Arctic.

**Acknowledgments**
We are grateful to Nathan J Pamperin of ADF&G for coordinating the data-sharing agreement and for his continued assistance with survey-data documentation throughout this study. We also thank Laureen McLaughlin for the final English revision.

**Author's Declaration**
All authors have reviewed and approved the final version of the manuscript. Each author has made significant contributions to the work, and all individuals meeting the criteria for authorship have been included.

**Author's contribution**
Ghazaleh Serati, Samuel Foucher and Jérôme Théau conceived the ideas and designed the methodology; Ghazaleh Serati, Samuel Foucher, and Jérôme Théau analysed the data; Ghazaleh Serati led the writing of the manuscript; and Ghazaleh Serati, Samuel Foucher, and Jérôme Théau contributed critically to the drafts. All authors gave final approval for publication.

**Data availability statement**
The original contributions presented in the study are included in the article and in supplementary files, further inquiries can be directed to the corresponding author/s.

# Appendices

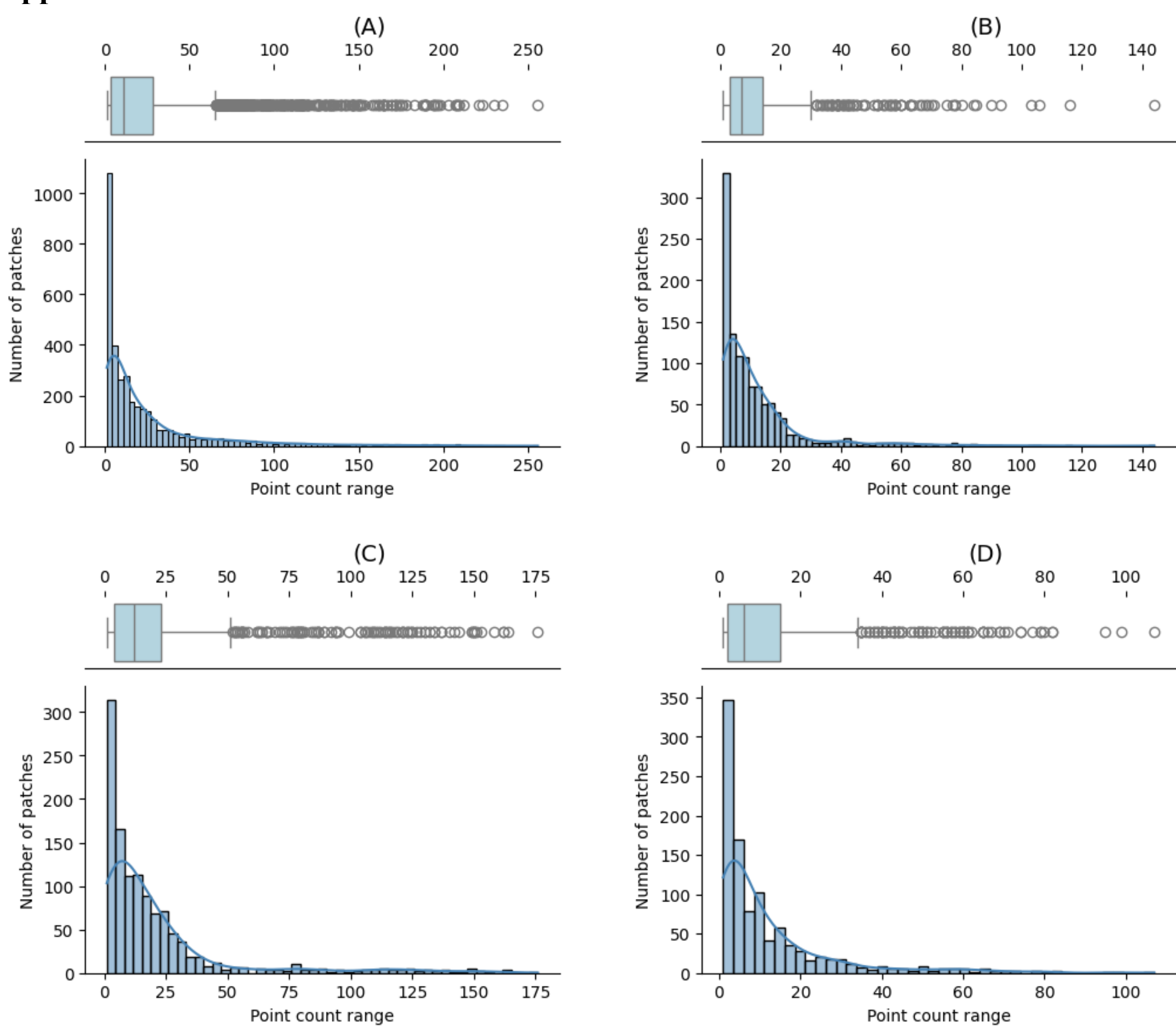

**Appendix 1 :** Distribution of annotated caribou points per patch across (A) training, (B) validation, (C) Test-2017, and (D) Test-2019 datasets.
The histograms illustrate the frequency of patches containing different numbers of caribou, highlighting the range of densities encountered across the datasets.
The box plots above each histogram provide a visual summary of density variability, showing how most patches have low to moderate counts, while a smaller number contain much denser aggregations.
The training set (A) exhibits the highest variability in caribou density, from sparse to very dense clusters (up to about 250 caribou per patch). In contrast, the validation and test sets (B–D) show fewer extreme values, reflecting smaller sample sizes and a lower proportion of highly dense patches.
This density variability across the subsets ensures that the model is exposed to the full range of caribou aggregation patterns, supporting better generalization to different ecological conditions

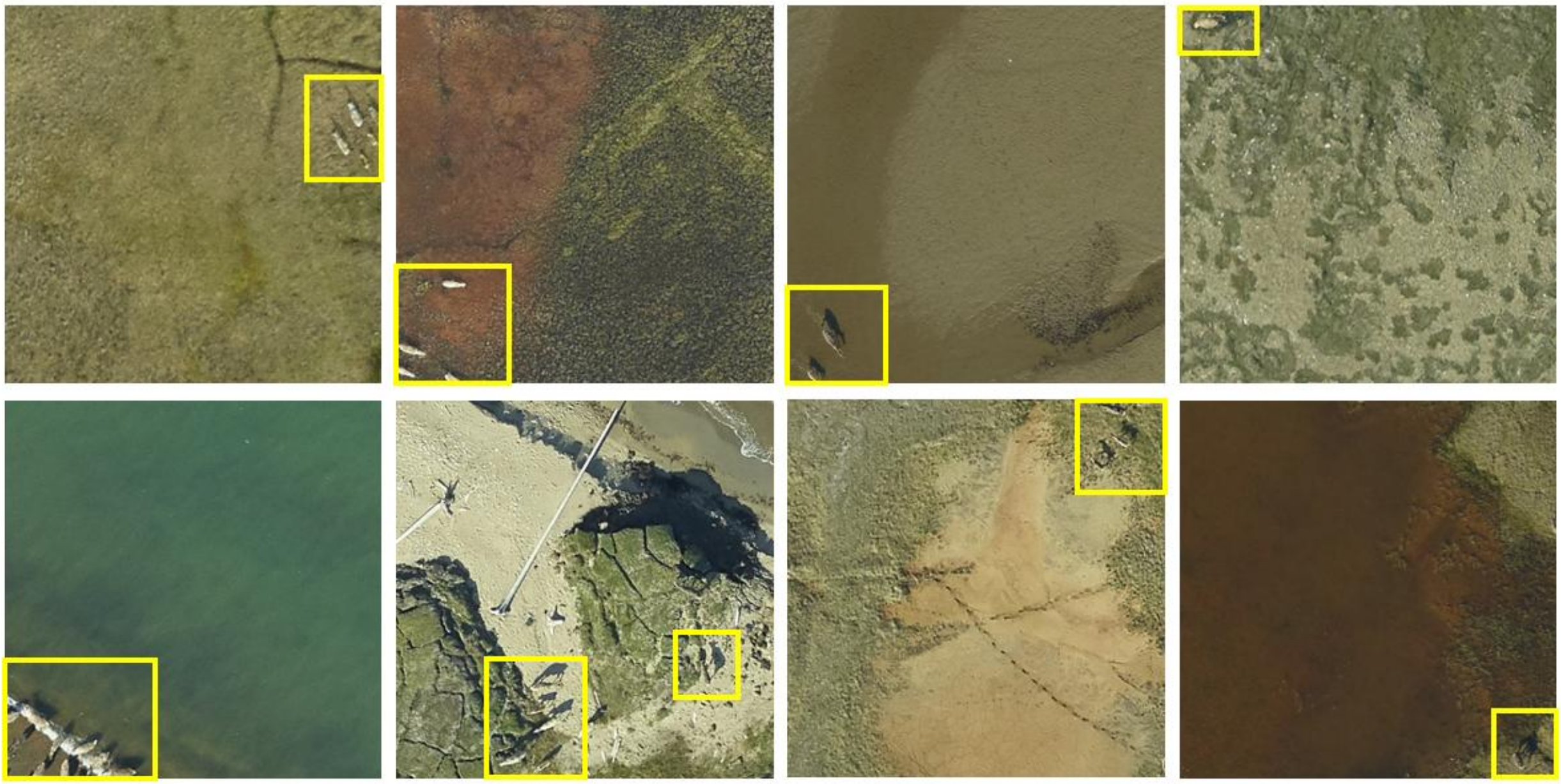

**Appendix 2 :** Samples of false negative (FN) patches detected by a Patch-based Pretrain Network initialized from ImageNet weights (PPN-ImageNet), including patches with one or more caribou. Top row: Examples from Test-2017; Bottom row: Examples from Test-2019.

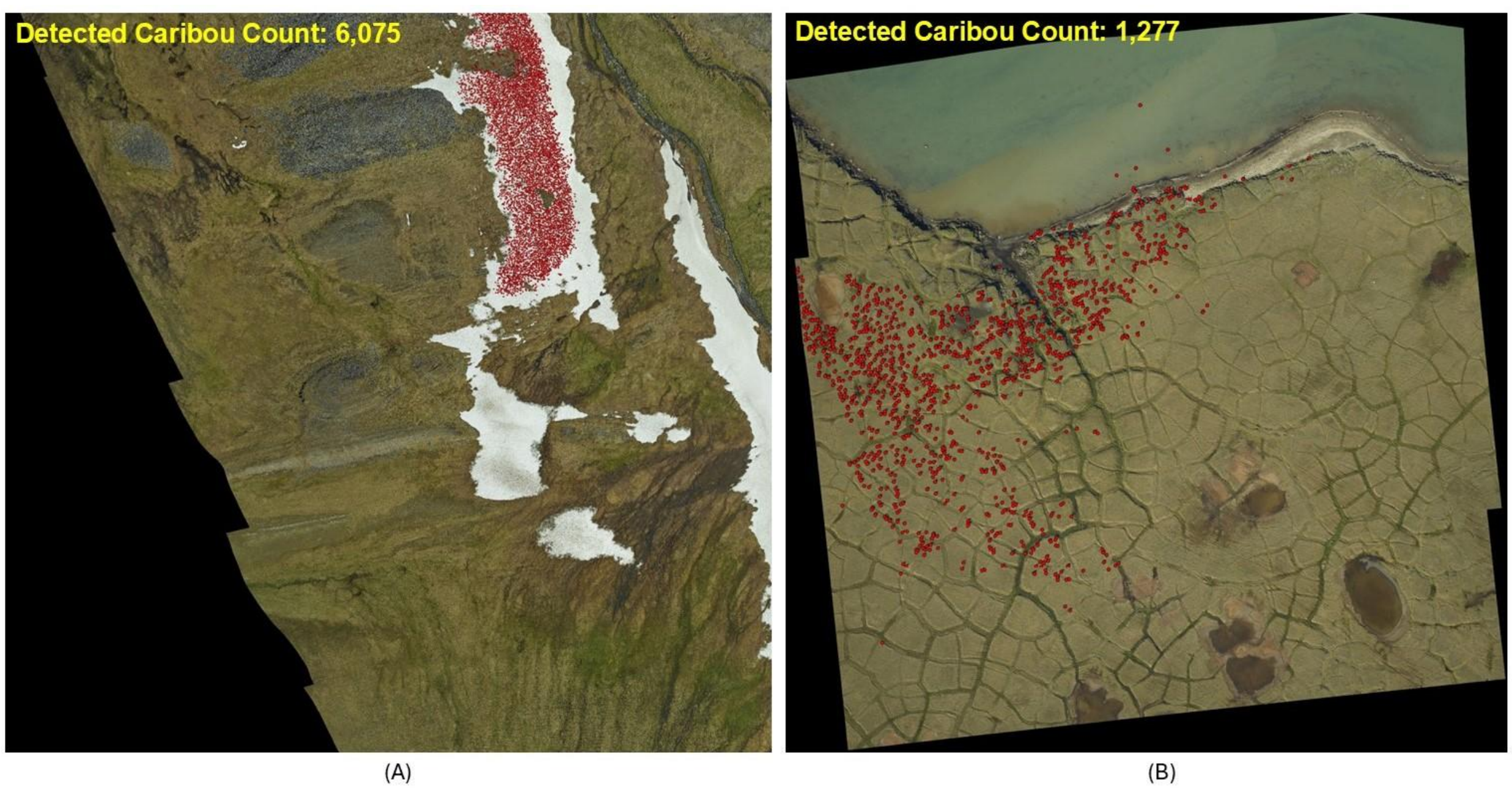

**Appendix 3 :** Examples of full-size image detections represented with red points and their corresponding counts: (A) Porcupine Caribou Herd associated with Test-2017; Area of Interest (AOI): 450,922.5 m² (≈ 45 ha); (B) Central arctic Herd associated with Test-2019; AOI: 91,320 m2 (≈ 9.1 ha).

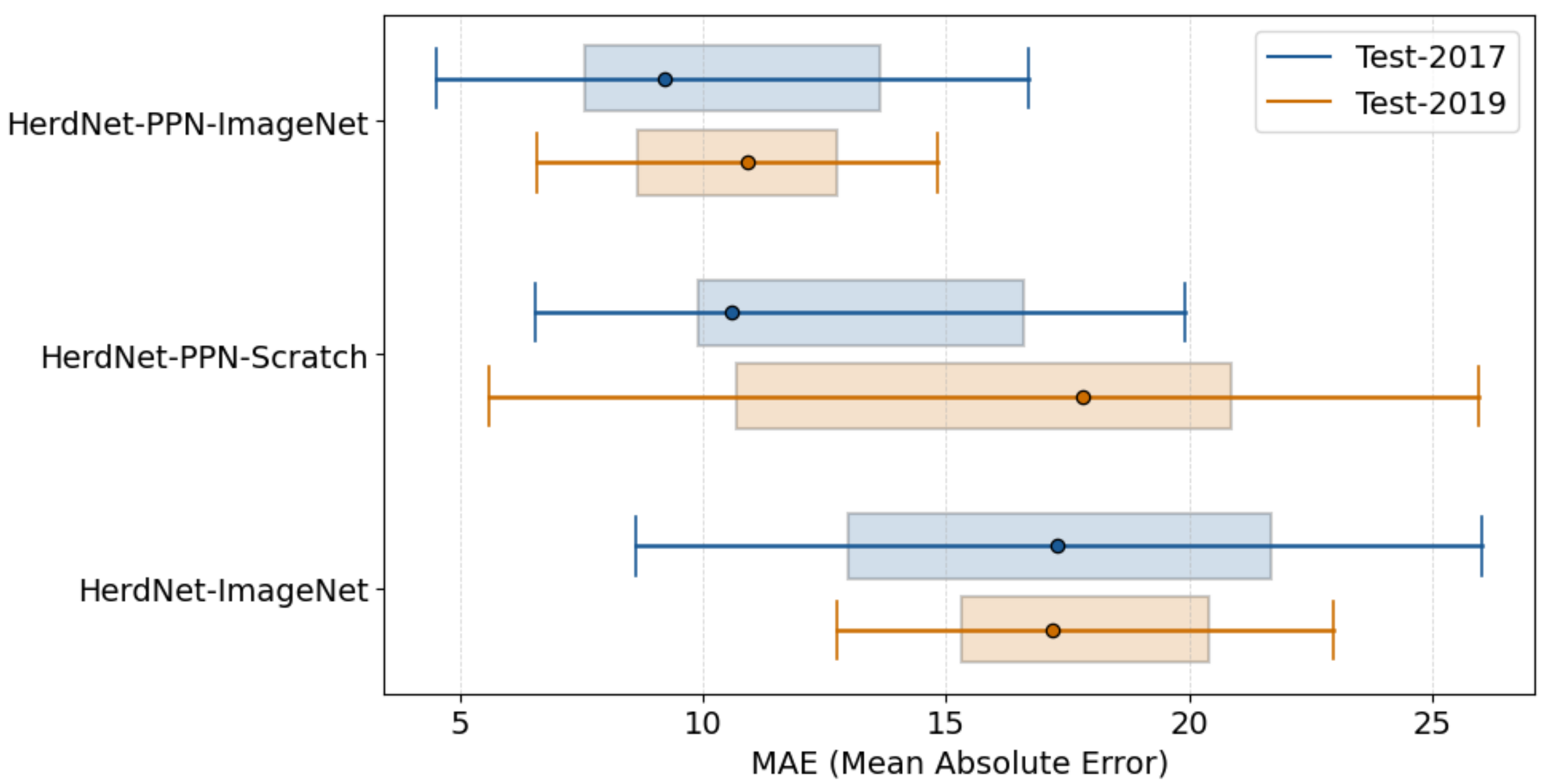

**Appendix 4 :** Mean Absolute Error (MAE) box plots of the caribou count on the full-size images on the two test sets. The MAE values of the three networks are marked with a point, and the vertical error bars indicate the corresponding 95% bootstrap confidence intervals. Boxes depict the 50% confidence interval for visual clarity.